\documentclass[11pt,letterpaper]{article}
\usepackage{naaclhlt2013}
\usepackage{times}
\usepackage{latexsym}
\usepackage{covington}
\usepackage{color}

\def\ul#1{$\underline{\smash{\hbox{#1}}}$}

\newenvironment{packed_item}
{\begin{itemize}
\setlength{\itemsep}{1pt}
\setlength{\parskip}{0pt}
\setlength{\parsep}{0pt}}
{\end{itemize}}

\title{SemEval-2013 Task 4: Free Paraphrases of Noun Compounds}

\author{
{\bf Iris Hendrickx}\\Radboud University Nijmegen \&\\Universidade de Lisboa\\{\small {\tt iris@clul.ul.pt}}\\
{\bf Preslav Nakov}\\QCRI, Qatar Foundation\\{\small {\tt pnakov@qf.org.qa}}\\
{\bf Stan Szpakowicz}\\ University of Ottawa \&\\Polish Academy of Sciences\\{\small {\tt szpak@eecs.uottawa.ca}}\\
\And
{\bf Zornitsa Kozareva}\\University of Southern California\\{\small {\tt kozareva@isi.edu}}\\~\\
{\bf Diarmuid \'{O} S\'{e}aghdha}\\University of Cambridge\\{\small {\tt do242@cam.ac.uk}}\\
{\bf Tony Veale}\\University College Dublin\\{\small {\tt tony.veale@ucd.ie}}
}

\date{}

\begin{document}
\maketitle
\begin{abstract} 
 In this paper, we describe SemEval-2013 Task 4: the definition, the data, the evaluation and the results.
 The task is to capture some of the meaning of English noun compounds via paraphrasing. Given a two-word noun compound, the participating system is asked to produce an explicitly ranked list of its free-form paraphrases. The list is automatically compared and evaluated against a similarly ranked list of paraphrases proposed by human annotators, recruited and managed through Amazon's Mechanical Turk. The comparison of raw paraphrases is sensitive to syntactic and morphological variation. The ``gold'' ranking is based on the relative popularity of paraphrases among annotators. To make the ranking more reliable, highly similar paraphrases are grouped, so as to downplay superficial differences in syntax and morphology. Three systems participated in the task. They all beat a simple baseline on one of the two evaluation measures, but not on both measures. This shows that the task is difficult.
\end{abstract}

\section{Introduction}

 A noun compound (NC) is a sequence of nouns which act as a single noun \cite{Downing:1977}, as in these examples: \emph{colon cancer}, \emph{suppressor protein}, \emph{tumor suppressor protein}, \emph{colon cancer tumor suppressor protein}, etc. This type of compounding is highly productive in English. NCs comprise 3.9\% and 2.6\% of all tokens in the Reuters corpus and the British National Corpus (BNC), respectively \cite{baldwin-tanaka:04}.

 The frequency spectrum of compound types follows a Zipfian distribution \cite{OSeaghdha:PhD}, so many NC tokens belong to a ``long tail" of low-frequency types. More than half of the two-noun types in the BNC occur exactly once \cite{Kim:Baldwin:06}. Their high frequency and high productivity make robust NC interpretation an important goal for broad-coverage semantic processing of English texts. Systems which ignore NCs may give up on salient information about the semantic relationships implicit in a text. Compositional interpretation is also the only way to achieve broad NC coverage, because it is not feasible to list in a lexicon all compounds which one is likely to encounter. Even for relatively frequent NCs occurring 10 times or more in the BNC, static English dictionaries provide only 27\% coverage \cite{Tanaka:Baldwin:2003}.

In many natural language processing applications it is important to understand the syntax and semantics of NCs. NCs often are structurally similar, but have very different meaning. Consider \emph{caffeine headache} and \emph{ice-cream headache}: a lack of caffeine causes the former, an excess of ice-cream -- the latter. Different interpretations can lead to different inferences, query expansion, paraphrases, translations, and so on. A question answering system may have to determine whether \emph{protein acting as a tumor suppressor} is an accurate paraphrase for \emph{tumor suppressor protein}. An information extraction system might need to decide whether \emph{neck vein thrombosis} and \emph{neck thrombosis} can co-refer in the same document. A machine translation system might paraphrase the unknown compound \emph{WTO Geneva headquarters} as
\emph{WTO headquarters located in Geneva}.

Research on the automatic interpretation of NCs has focused mainly on common two-word NCs. The usual task is to classify the semantic relation underlying a compound with either one of a small number of predefined relation labels or a paraphrase from an open vocabulary. Examples of the former take on classification include \cite{moldovan-EtAl:2004:HLTNAACL,Girju:07,OSeaghdha:Copestake:08,Tratz:Hovy:10}. Examples of the latter include \cite{Nakov:08a,Nakov:08MT,Nakov:Hearst:08,Butnariu:Veale:08} and a previous NC paraphrasing task at SemEval-2010 \cite{butnariu-EtAl:2010:SemEval}, upon which the task described here builds.

The assumption of a small inventory of predefined relations has some advantages -- parsimony and generalization -- but at the same time there are limitations on expressivity and coverage. For example, the NCs \emph{headache pills} and \emph{fertility pills} would be assigned the same semantic relation (\emph{PURPOSE}) in most inventories, but their relational semantics are quite different \cite{Downing:1977}. Furthermore, the definitions given by human subjects can involve rich and specific meanings. For example, \newcite{Downing:1977} reports that a subject defined the NC \emph{oil bowl} as ``the bowl into which the oil in the engine is drained during an oil change'', compared to which a minimal interpretation \emph{bowl for oil} seems very reductive. In view of such arguments, linguists such as \newcite{Downing:1977}, \newcite{Ryder:94} and \newcite{Coulson:01} have argued for a fine-grained, essentially open-ended space of interpretations.

The idea of working with fine-grained paraphrases for NC semantics has recently grown in popularity among NLP researchers \cite{Butnariu:Veale:08,Nakov:Hearst:08,Nakov:08MT}. Task 9 at SemEval-2010 \cite{butnariu-EtAl:2010:SemEval} was devoted to this methodology. In that previous work, the paraphrases provided by human subjects were required to fit a restrictive template admitting only verbs and prepositions occurring between the NC's constituent nouns. Annotators recruited through Amazon Mechanical Turk were asked to provide paraphrases for the dataset of NCs. The gold standard for each NC was the ranked list of paraphrases given by the annotators; this reflects the idea that a compound's meaning can be described in different ways, at different levels of granularity and capturing different interpretations in the case of ambiguity.

For example, a \emph{plastic saw} could be a \emph{saw made of plastic} or a \emph{saw for cutting plastic}. Systems participating in the task were given the set of attested paraphrases for each NC, and evaluated according to how well they could reproduce the humans' ranking.

The design of this task, SemEval-2013 Task 4, is informed by previous work on compound annotation and interpretation. It is also influenced by similar initiatives, such as the English Lexical Substitution task at SemEval-2007 \cite{mccarthy:2007:SemEval-2007}, and by various evaluation exercises in the fields of paraphrasing and machine translation. We build on SemEval-2010 Task 9, extending the task's flexibility in a number of ways. The restrictions on the form of annotators' paraphrases was relaxed, giving us a rich dataset of close-to-freeform paraphrases (Section \ref{sec:data}). Rather than ranking a set of attested paraphrases, systems must now both generate and rank their paraphrases; the task they perform is essentially the same as what the annotators were asked to do. This new setup required us to innovate in terms of evaluation measures (Section \ref{sec:eval}).

We anticipate that the dataset and task will be of broad interest among those who study lexical semantics. We believe that the overall progress in the field will significantly benefit from a public-domain set of free-style NC paraphrases. That is why our primary objective is the challenging endeavour of preparing and releasing such a dataset to the research community. The common evaluation task which we establish will also enable researchers to compare their algorithms and their empirical results.

\section{Task description}\label{sec:task}

This is an English NC interpretation task, which explores the idea of interpreting the semantics of NCs via free paraphrases. Given a noun-noun compound such as \emph{air filter}, the participating systems are asked to produce an explicitly ranked list of free paraphrases, as in the following example:

\begin{examples}
\item[]
1	filter for air \\ 
2	filter of air \\ 
3	filter that cleans the air \\ 
4	filter which makes air healthier\\ 
5	a filter that removes impurities from the air\\ 
\ldots
\end{examples}

Such a list is then automatically compared and evaluated against a similarly ranked list of paraphrases proposed by human annotators, recruited and managed via Amazon's Mechanical Turk. The comparison of raw paraphrases is sensitive to syntactic and morphological variation. The ranking of paraphrases is based on their relative popularity among different annotators. To make the ranking more reliable, highly similar paraphrases are grouped so as to downplay superficial differences in syntax and morphology.

\section{Data collection}\label{sec:data}

We used Amazon's \emph{Mechanical Turk} service to collect diverse paraphrases for a range of ``gold-standard'' NCs.\footnote{Since the annotation on Mechanical Turk was going slowly, we also recruited four other annotators to do the same work, following exactly the same instructions.} We paid the workers a small fee (\$0.10) per compound, for which they were asked to provide five paraphrases. Each paraphrase should contain the two nouns of the compound (in singular or plural inflectional forms, but not in another derivational form), an intermediate non-empty linking phrase and optional preceding or following terms. The paraphrasing terms could have any part of speech, so long as the resulting paraphrase was a well-formed noun phrase headed by the NC's head.

We gave the workers feedback during data collection if they appeared to have misunderstood the nature of the task. Once raw paraphrases had been collected from all workers, we collated them into a spreadsheet, and we merged identical paraphrases in order to calculate their overall frequencies. Ill-formed paraphrases -- those violating the syntactic restrictions described above -- were manually removed following a consensus decision-making procedure; every paraphrase was checked by at least two task organizers. We did not require that the paraphrases be semantically felicitous, but we performed minor edits on the remaining paraphrases if they contained obvious typos.

The remaining well-formed paraphrases were sorted by frequency separately for each NC. The most frequent paraphrases for a compound are assigned the highest rank 0, those with the next-highest frequency are given a rank of 1, and so on.

Paraphrases with a frequency of 1 -- proposed for a given NC by only one annotator -- always occupy the lowest rank on the list for that compound.

We used 174+181 noun-noun compounds from the NC dataset of \newcite{OSeaghdha:07}.
The trial dataset, which we initially released to the participants, consisted of 4,255 human paraphrases for 174 noun-noun pairs;
this dataset was also the training dataset.
The test dataset comprised paraphrases for 181 noun-noun pairs.
The ``gold standard'' contained 9,706 paraphrases of which 8,216 were unique for those 181 NCs.
Further statistics on the datasets are presented in Table \ref{stat-table}.

\begin{table}[t]
\begin{center}
\begin{tabular}{lcc}
& {\bf Total} & {\bf Min / Max / Avg} \\[1ex]
\multicolumn{2}{l}{\ul{Trial/Train (174 NCs)}} & \\
paraphrases & 6,069 & 1 / 287 / 34.9 \\
unique paraphrases & 4,255 & 1 / 105 / 24.5 \\[1ex]
\ul{Test (181 NCs)} & & \\
paraphrases & 9,706 & 24 / 99 / 53.6 \\
unique paraphrases & 8,216 & 21 / 80 / 45.4 \\
\end{tabular}
\end{center}
\caption{Statistics of the trial and test datasets: the total number of paraphrases with and without duplicates, and the minimum / maximum / average per noun compound.} 
\label{stat-table}
\end{table}

Compared with the data collected for the SemEval-2010 Task 9 on the interpretation of noun compounds, the data collected for this new task have a far greater range of variety and richness. For example, the following (selected) paraphrases for \emph{work area} vary from parsimonious to expansive:
\begin{packed_item}
\item area for work
\item area of work
\item area where work is done
\item area where work is performed
\item \ldots
\item an area cordoned off for persons responsible for work
\item an area where construction work is carried out
\item an area where work is accomplished and done
\item area where work is conducted
\item office area assigned as a work space
\item \ldots
\end{packed_item}

\section{Scoring}\label{sec:eval}

 Noun compounding is a generative aspect of language, but so too is the process of NC interpretation: human speakers typically generate a range of possible interpretations for a given compound, each emphasizing a different aspect of the relationship between the nouns. Our evaluation framework reflects the belief that there is rarely a single right answer for a given noun-noun pairing. Participating systems are thus expected to demonstrate some generativity of their own, and are scored not just on the accuracy of individual interpretations, but on the overall breadth of their output.

For evaluation, we provided a scorer implemented, for good portability, as a Java class. For each noun compound to be evaluated, the scorer compares a list of system-suggested paraphrases against a ``gold-standard'' reference list, compiled and rank-ordered from the paraphrases suggested by our human annotators. The score assigned to each system is the mean of the system's performance across all test compounds. Note that the scorer removes all determiners from both the reference and the test paraphrases, so a system is neither punished for not reproducing a determiner or rewarded for producing the same determiners.

The scorer can match words identically or non-identically. A match of two identical words $W_{gold}$ and $W_{test}$ earns a score of 1.0. There is a partial score of $(2~|P|~/~(|PW_{gold}|$ + $|PW_{test}|))^2$ for a match of two words $PW_{gold}$ and $PW_{test}$ that are not identical but share a common prefix $P$, $|P|>2$, e.g., $wmatch(\emph{cutting}, \emph{cuts})$ = $(6/11)^2$ = 0.297.

Two $n$-grams $N_{gold}$ = [$GW_1$, \ldots, $GW_n$] and $N_{test}$ = [$TW_1$, \ldots, $TW_n$] can be matched if $wmatch(GW_i,~TW_i) > 0$ for all $i$ in $1 .. n$. The score assigned to the match of these two $n$-grams is then $\sum_i wmatch(GW_i,~TW_i)$. For every $n$-gram $N_{test}$ = [$TW_1$, \ldots, $TW_n$] in a system-generated paraphrase, the scorer finds a matching $n$-gram $N_{gold}$ = [$GW_1$, \ldots, $GW_n$] in the reference paraphrase $Para_{gold}$ which maximizes this sum.

The overall $n$-gram overlap score for a reference paraphrase $Para_{gold}$ and a system-generated paraphrase $Para_{test}$ is the sum of the score calculated for all $n$-grams in $Para_{test}$, where $n$ ranges from 1 to the size of $Para_{test}$.

This overall score is then normalized by dividing by the maximum value among the $n$-gram overlap score for $Para_{gold}$ compared with itself and the $n$-gram overlap score for $Para_{test}$ compared with itself. This normalization step produces a paraphrase match score in the range [0.0 \mbox{--} 1.0]. It punishes a paraphrase $Para_{test}$ for both over-generating (containing more words than are found in $Para_{gold}$) and under-generating (containing fewer words than are found in $Para_{gold}$). In other words, $Para_{test}$ should ideally reproduce everything in $Para_{gold}$, and nothing more or less.

The reference paraphrases in the ``gold standard'' are ordered by rank; the highest rank is assigned to the paraphrases which human judges suggested most often. The rank of a reference paraphrase matters because a good participating system will aim to reproduce the top-ranked ``gold-standard'' paraphrases as produced by human judges. The scorer assigns a multiplier of $R/(R + n)$ to reference paraphrases at rank $n$; this multiplier asymptotically approaches 0 for the higher values of $n$ of ever lower-ranked paraphrases. We choose a default setting of $R=8$,  so that a reference paraphrase at rank 0 (the highest rank) has a multiplier of 1, while a reference paraphrase at rank 5 has a multiplier of $8/13$ = 0.615.

When a system-generated paraphrase $Para_{test}$ is matched with a reference paraphrase $Para_{gold}$, their normalized $n$-gram overlap score is scaled by the rank multiplier attaching to the rank of $Para_{gold}$ relative to the other reference paraphrases provided by human judges. The scorer automatically chooses the reference paraphrase $Para_{gold}$ for a test paraphrase $Para_{test}$ so as to maximize this product of normalized $n$-gram overlap score and rank multiplier.

The overall score assigned to each system for a specific compound is calculated in two different ways: using \emph{isomorphic matching} of suggested paraphrases to the ``gold-standard's'' reference paraphrases (on a \emph{one-to-one} basis); and using \emph{non-isomorphic matching} of system's paraphrases to the ``gold-standard's'' reference paraphrases (in a potentially \emph{many-to-one} mapping).

\emph{Isomorphic matching} rewards both precision and recall. It rewards a system for accurately reproducing the paraphrases suggested by human judges, and for reproducing as many of these as it can, and in much the same order.

In isomorphic mode, system's paraphrases are matched 1-to-1 with reference paraphrases on a first-come first-matched basis, so ordering can be crucial.

\emph{Non-isomorphic} matching rewards only precision. It rewards a system for accurately reproducing the top-ranked human paraphrases in the ``gold standard''. A system will achieve a higher score in a non-isomorphic match if it reproduces the top-ranked human paraphrases as opposed to lower-ranked human paraphrases. The ordering of system's paraphrases is thus not important in non-isomorphic matching.

Each system is evaluated using the scorer in both modes, \emph{isomorphic} and \emph{non-isomorphic}. Systems which aim only for precision should score highly on non-isomorphic match mode, but poorly in isomorphic match mode. Systems which aim for precision \emph{and} recall will face a more substantial challenge, likely reflected in their scores.

\paragraph{A na\"{i}ve baseline}~\\
We decided to allow preposition-only paraphrases, which are abundant in the paraphrases suggested by human judges in the crowdsourcing Mechanical Turk collection process. This abundance means that the top-ranked paraphrase for a given compound is often a preposition-only phrase, or one of a small number of very popular paraphrases such as \emph{used for} or \emph{used in}. It is thus straightforward to build a na\"{i}ve baseline generator which we can expect to score reasonably on this task, at least in \emph{non-isomorphic matching} mode. For each test compound \emph{M H}, the baseline system generates the following paraphrases, in this precise order: \emph{H of M}, \emph{H in M}, \emph{H for M}, \emph{H with M}, \emph{H on M}, \emph{H about M}, \emph{H has M}, \emph{H to M}, \emph{H used for M}, \emph{H used in M}.

This na\"{i}ve baseline is truly unsophisticated. No attempt is made to order paraphrases by their corpus frequencies or by their frequencies in the training data. The same sequence of paraphrases is generated for each and every test compound.

\section{Results}

Three teams participated in the challenge, and all their systems were supervised.
The MELODI system relied on semantic vector space model built from the UKWAC corpus (window-based, 5 words). It used only the features of the right-hand head noun to train a maximum entropy classifier.

The IIITH system used the probabilities of the preposition co-occurring with a relation to identify the class of the noun compound. To collect statistics, it used Google $n$-grams, BNC and ANC.

The SFS system extracted templates and fillers from the training data, which it then combined with a four-gram language model and a MaxEnt reranker. To find similar compounds, they used Lin's WordNet similarity. They further used statistics from the English Gigaword and the Google $n$-grams.

Table~\ref{tbl:results} shows the performance of the participating systems, SFS, IIITH and MELODI, and the na\"{i}ve baseline. The baseline shows that it is relatively easy to achieve a moderately good score in non-isomorphic match mode by generating a fixed set of paraphrases which are both common and generic: two of the three participating systems, SFS and IIITH, under-perform the na\"{i}ve baseline in non-isomorphic match mode, but outperform it in isomorphic mode. The only system to surpass this baseline in non-isomorphic match mode is the MELODI system; yet, it under-performs against the same baseline in isomorphic match mode. No participating team submitted a system which would outperform the na\"{i}ve baseline in both modes.

\begin{table}
\begin{center}
\begin{tabular}{|l|cc|}
\hline \bf Team & isomorphic & non-isomorphic \\ \hline
SFS & 23.1 & 17.9\\
IIITH & 23.1 & 25.8\\
MELODI-Primary & 13.0 & 54.8\\
MELODI-Contrast &13.6 & 53.6\\
\emph{Naive Baseline} & \emph{13.8} & \emph{40.6}\\
\hline
\end{tabular}
\end{center}
\caption{\label{tbl:results} Results for the participating systems; the baseline outputs the same paraphrases for all compounds.}
\end{table}

\section{Conclusions}

The conclusions we draw from the experience of organizing the task are mixed. Participation was reasonable but not large, suggesting that NC paraphrasing remains a niche interest -- though we believe it deserves more attention among the broader lexical semantics community and hope that the availability of our freeform paraphrase dataset will attract a wider audience in the future.

We also observed a varied response from our annotators in terms of embracing their freedom to generate complex and rich paraphrases; there are many possible reasons for this including laziness, time pressure and the fact that short paraphrases are often very appropriate paraphrases. The results obtained by our participants were also modest, demonstrating that compound paraphrasing is both a difficult task and a novel one that has not yet been ``solved''.




\section*{Acknowledgments}

This work has partially supported by a small but effective grant from Amazon; the credit allowed us to hire sufficiently many Turkers -- thanks! And a thank-you to our additional annotators Dave Carter, Chris Fournier and Colette Joubarne for their complete sets of paraphrases of the noun compounds in the test data.


\end{document}